\documentclass[runningheads]{llncs}

\usepackage{eccv}

\usepackage{eccvabbrv}

\usepackage{graphicx}
\usepackage{booktabs}

\usepackage[accsupp]{axessibility}  

\usepackage{hyperref}

\usepackage{orcidlink}

\usepackage{multirow} 
\usepackage{algorithm}
\usepackage{algorithmic}
\usepackage{colortbl}
\usepackage{amssymb}

\usepackage{smil}
\def\shortname{GAST\xspace}

\begin{document}

\title{Geometry-Aware Spatio-Temporal Context Modeling for 4D Occupancy Forecasting} 

\titlerunning{Geometry-Aware Spatio-Temporal Context Modeling for 4D Occ Forecasting}

\author{Sitao Chen\inst{1} \and Zhuangwei Zhuang\inst{1} \and Hui Luo\inst{2}$^\star$ \and \\ Qingyao Wu\inst{1} \and Mingkui Tan\inst{1}\thanks{Corresponding authors.}}

\authorrunning{S. Chen et al.}

\institute{South China University of Technology, Guangzhou, China \and Institute of Optics and Electronics, CAS, Chengdu, China 
\email{chensitao27@gmail.com, mingkuitan@scut.edu.cn}}

\maketitle

\begin{abstract}
4D occupancy forecasting models the spatio-temporal evolution of 3D scenes and is crucial for autonomous driving, especially for corner-case simulation. Existing methods often rely on discrete tokenization followed by autoregressive prediction, yet struggle with geometric distortion in static structures and inconsistent temporal coherence over the forecasting horizon. In this work, we propose a Geometry-Aware Spatio-Temporal context modeling method (\shortname) for 4D occupancy forecasting, built upon progressive explicit-implicit generation and dual-path spatio-temporal modeling. Specifically, the generation module produces per-frame occupancy with high geometric fidelity and semantic plausibility through pose-driven warping, motion-aware feature modulation, and attention-based feature refinement. Subsequently, the spatio-temporal module enhances spatial consistency through global context aggregation while capturing scene evolution through temporal dynamics extraction. This unified design enables joint optimization of historical reconstruction and future forecasting in an end-to-end manner. Extensive experiments on Occ3D-nuScenes demonstrate the superiority of our method, outperforming the state-of-the-art by 7.67\% in mIoU and 6.44\% in IoU with a $2.84\times$ speedup, while maintaining strong performance in long-term forecasting. Our source code is publicly available at \url{https://github.com/chenst27/GAST}.


\end{abstract}    
\section{Introduction}
3D occupancy represents driving scenes through dense voxel grids with geometric occupancy and semantic labels, offering a holistic and fine-grained 3D representation~\cite{zhuang2024robust, wei2023surroundocc, wang2023openoccupancy, behley2019semantickitti}. Extending this into the temporal domain, 4D occupancy forecasting~\cite{zheng2024occworld, liao2025i2, gu2024dome, shi2025come} predicts future 3D scene states from historical observations to model spatio-temporal evolution, supporting critical downstream applications, such as long-horizon motion planning~\cite{zheng2025world4drive, li2024enhancing}, simulation of corner-case scenarios~\cite{bian2024dynamiccity, wang2024occsora}, and synthesis of high-fidelity data~\cite{yang2025drivearena, yan2025drivingsphere}. As a result, 4D occupancy forecasting has emerged as a fundamental task for modern autonomous driving systems.

\begin{figure}[t]
    \centering
    \includegraphics[width=1.0\linewidth]{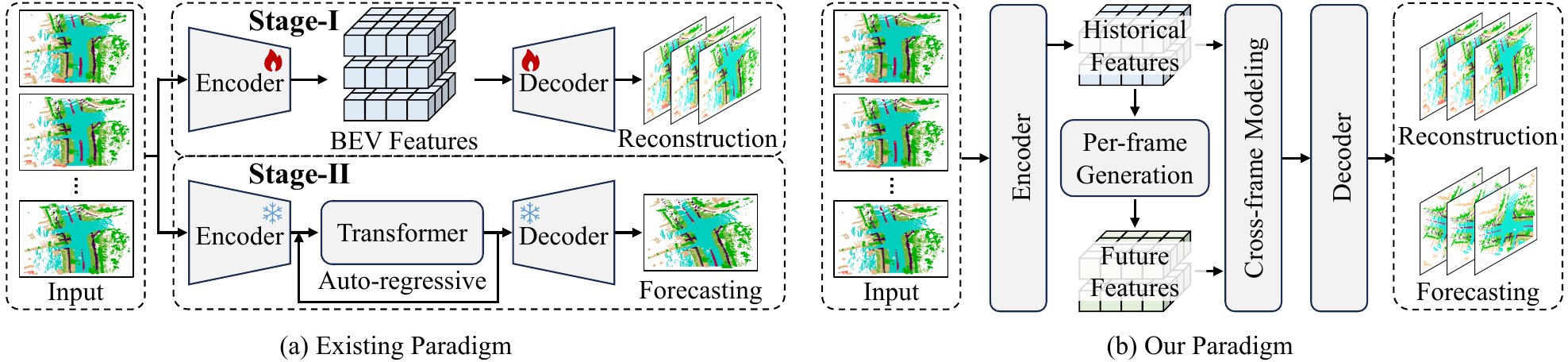}
    \caption{Comparison with the existing paradigm~\cite{zheng2024occworld, liao2025i2}. It adopts a two-stage pipeline that encodes occupancy into discrete tokens and autoregressively generates future tokens. In contrast, our end-to-end framework directly models spatio-temporal representations in continuous BEV space via per-frame generation and cross-frame modeling.}
    \label{fig:paradigm_cmp}
\end{figure}

Driven by advances in generative AI~\cite{brown2020language}, autoregressive-based occupancy world models~\cite{zheng2024occworld, liao2025i2, wei2024occllama, xu2025occ} have become the dominant paradigm for 4D occupancy forecasting (See Figure~\ref{fig:paradigm_cmp}). These methods typically compress 3D semantic occupancy into discrete tokens via VQ-VAE~\cite{van2017neural}, then autoregressively generate future tokens for decoding into occupancy grids. However, they suffer from three major limitations. First, the lack of explicit geometric constraints on static scene structures leads to shape distortion or positional drift (\eg, roads, buildings) during long-term prediction. Second, the causal, step-by-step nature of autoregressive generation hinders the modeling of global spatial context across timestamps, compromising spatio-temporal consistency. Third, the two-stage training pipeline (\ie, separating tokenization from sequence modeling) leaves the predictor inherently constrained by the tokenizer's representational capacity. Addressing these limitations requires a 4D occupancy forecasting framework that jointly ensures geometric consistency of static structures, plausible semantic evolution of dynamic scenes, and cross-frame coherence over the prediction horizon.

In this work, we propose a Geometry-Aware Spatio-Temporal context modeling framework (\shortname) that integrates progressive explicit-implicit generation and dual-path spatio-temporal modeling. Given historical occupancy and ego-poses, we first encode compact bird's-eye-view (BEV) features and predict future ego-poses. The generation module then propagates the current BEV structure into future frames via pose-driven rigid transformation, thereby enforcing geometric consistency for static elements. The transformed futures are refined through two implicit stages: motion-aware feature modulation injects sequential ego-dynamics via adaptive affine transformations, and attention-based feature refinement selectively fuses high-fidelity cues from the current observation to enable plausible dynamic evolution. Building on these per-frame representations, the dual-path spatio-temporal module processes spatial layout and temporal dynamics in parallel: a global spatial context aggregator unifies historical and future features in a shared world coordinate system to capture long-range structural dependencies, while a temporal dynamics extractor regularizes the semantic evolution through sequential encoding. Their outputs are fused to enforce cross-frame coherence in long-horizon predictions. Finally, our \shortname is trained end-to-end by jointly optimizing historical reconstruction and future forecasting, which eliminates complex two-stage pipelines and mitigates the error accumulation in autoregressive generation.
Our contributions are summarized as follows:

\begin{itemize}
    \item We propose a geometry-aware spatio-temporal context modeling method (\shortname) for 4D occupancy forecasting that constructs per-frame future representations through progressive explicit-implicit generation and enforces cross-frame consistency via dual-path spatio-temporal modeling, achieving geometric fidelity, semantic plausibility, and temporal coherence.
    \item We design a progressive explicit-implicit generation strategy that explicitly propagates current scene geometry to future frames via pose-driven warping and implicitly refines future representations through motion-aware modulation and attention-based refinement, enhancing both geometric consistency and plausible scene evolution.
    \item We introduce a dual-path spatio-temporal modeling module that processes spatial layout and temporal evolution in parallel through a global spatial aggregator and a sequential temporal extractor, then fuses their outputs to ensure long-horizon coherence. Extensive experimental results on Occ3D-nuScenes demonstrate the state-of-the-art performance of our method.
\end{itemize}
\section{Related Work}

\noindent\textbf{3D occupancy prediction.}
Semantic occupancy prediction estimates both the occupancy state and semantic label of every voxel in 3D space, offering a more comprehensive and expressive scene representation than object detection~\cite{yin2021center, li2024bevformer} or LiDAR segmentation~\cite{tan2024epmf, yan20222dpass}. Existing methods can be mainly categorized into camera-based, LiDAR-based, and multi-sensor fusion methods. Camera-based methods~\cite{huang2023tri, li2023voxformer, yu2024context, lu2025vishall3d} exploit rich appearance cues from images to infer 3D occupancy. CGFormer~\cite{yu2024context} introduces the context- and geometry-aware voxel Transformer to lift 2D features into 3D volumes. VisHall3D~\cite{lu2025vishall3d} proposes VisFrontierNet to trace the visible frontier and OcclusionMAE to hallucinate plausible geometries for occluded regions. LiDAR-based methods~\cite{yan2021sparse, yang2021semantic, mei2023ssc, wang2024voxel} leverage precise spatial measurements of point clouds for volumetric reconstruction. SSC-RS~\cite{mei2023ssc} captures multi-level semantic context and multi-scale geometric structures, fusing them through an adaptive representation module. VPNet~\cite{wang2024voxel} models semantic uncertainty by introducing confident voxels to represent diverse occupancy possibilities. Multi-sensor methods~\cite{pan2024co, wang2024occgen, zhuang2024robust, duan2025sdgocc} further exploit complementary strengths from the two modalities. Co-Occ~\cite{pan2024co} employs implicit volume rendering to fuse explicit LiDAR-camera features. OccGen~\cite{wang2024occgen} conditions occupancy generation on multi-modal features and progressively refines predictions. Despite their effectiveness in static perception, these methods fail to model scene dynamics and thus cannot capture temporal evolution.

\begin{figure*}[t]
    \centering
    \includegraphics[width=1.0\linewidth]{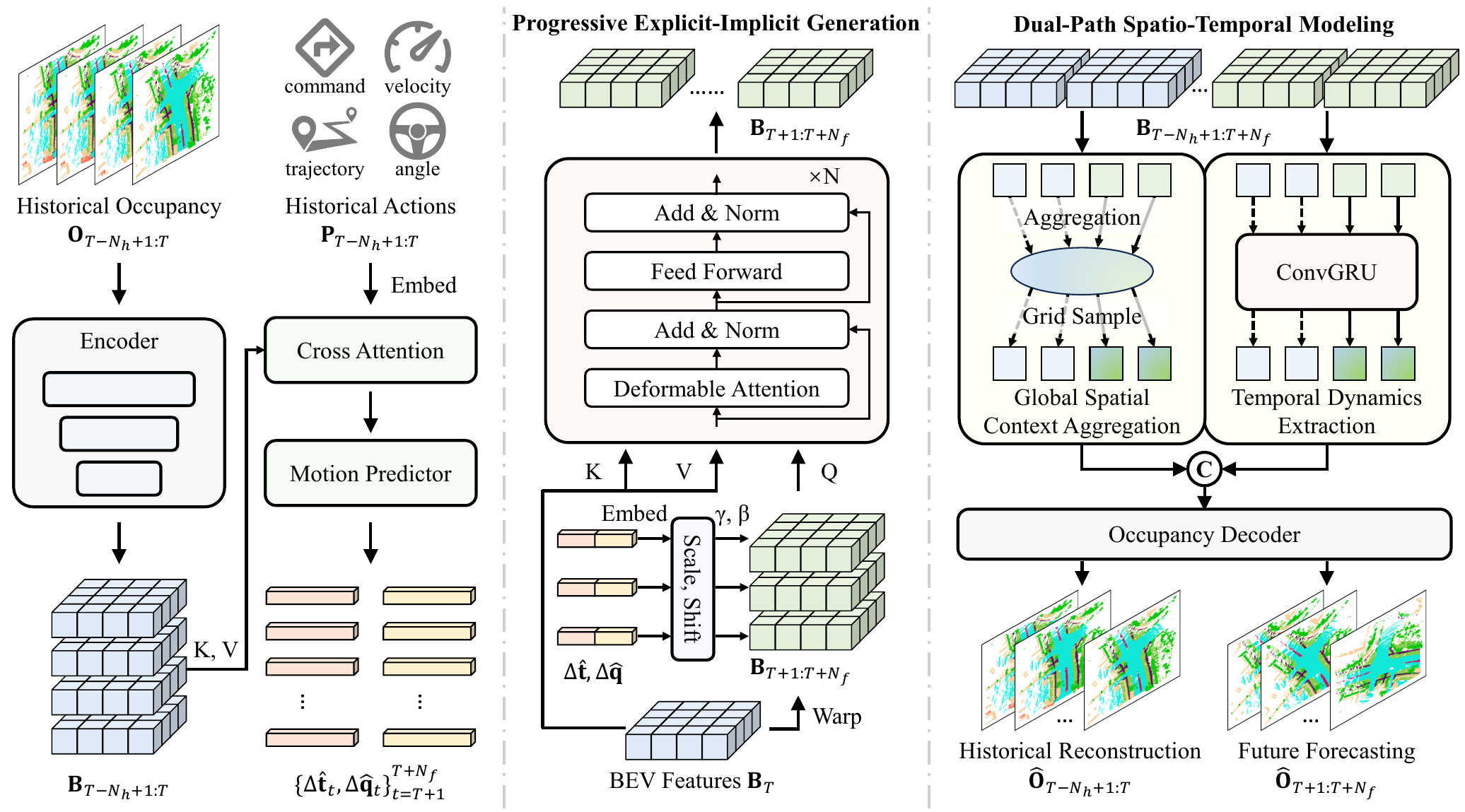}
    \caption{General architecture of \shortname.
    We first encode historical occupancy into BEV features and predict future ego-poses. The generation module outputs per-frame future features via geometric warping, feature modulation, and deformable cross-attention. The spatio-temporal module enhances global consistency through spatial context aggregation while modeling scene evolution via temporal extraction. Finally, an occupancy decoder jointly reconstructs past observations and forecasts future occupancy.
    }
    \label{fig:arch_overview}
\end{figure*}

\noindent\textbf{4D occupancy forecasting.}
4D occupancy forecasting aims to predict the future spatio-temporal evolution of 3D scenes in terms of both geometry and semantics. OccWorld~\cite{zheng2024occworld} pioneers the occupancy world model paradigm that first compresses 3D scenes into discrete tokens via a VQ-VAE tokenizer and then autoregressively generates future tokens with a generative transformer. RenderWorld~\cite{yan2025renderworld} proposes the Air-Masked VAE that decouples the encoding of air and non-air voxels to preserve the structural sparsity of 3D scenes. Occ-LLM~\cite{xu2025occ} introduces the Motion-Separation VAE that distinguishes between movable and immovable scene elements, and leverages a large language model for occupancy generation. OccLLaMA~\cite{wei2024occllama} unifies scene, language, and action tokens into a single vocabulary for joint occupancy-language-action generation. 
OccTENS~\cite{jin2025occtens} decomposes 4D occupancy forecasting into spatial scale-by-scale generation and temporal scene-by-scene prediction.
OccSora~\cite{wang2024occsora} adopts a 4D scene tokenizer to directly extract spatio-temporal representations from 4D occupancy and generates scenes using a diffusion transformer~\cite{peebles2023scalable}. DynamicCity~\cite{bian2024dynamiccity} further employs HexPlane~\cite{cao2023hexplane} as a compact 4D representation for both tokenization and generation, enabling efficient modeling of long-horizon scene dynamics. $I^{2}$-World~\cite{liao2025i2} decouples scene tokenization into intra-scene and inter-scene components, where the former captures fine-grained static details and the latter models dynamic scene evolution. DOME~\cite{gu2024dome} introduces a diffusion-based world model that compresses occupancy into a continuous latent space and generates future occupancy conditioned on historical observations. COME~\cite{shi2025come} further incorporates ControlNet~\cite{zhang2023adding} to convert diverse scene conditions into control features that guide the occupancy generation process. However, these methods lack explicit structural priors and cross-frame geometric alignment. In this paper, we propose a geometry-aware spatio-temporal modeling method that constructs future occupancy through progressive explicit-implicit generation and ensures long-horizon coherence via dual-path spatio-temporal modeling.
\section{Proposed Method}

\subsection{Overall Architecture}
As illustrated in Figure~\ref{fig:arch_overview}, our \shortname consists of an occupancy encoder (Section~\ref{sec:occ_encoder}), a progressive explicit-implicit generation module (Section~\ref{sec:ceig_module}), a dual-path spatio-temporal modeling module (Section~\ref{sec:dstm_module}) and an occupancy decoder (Section~\ref{sec:occ_decoder}). Given historical occupancy and ego-motion information, the encoder extracts compact BEV features and predicts future ego-poses. The generation module produces future BEV features through explicit geometric warping, motion-aware implicit modulation, and attention-based feature refinement. To ensure cross-frame coherence, the spatio-temporal module processes historical and initial future features in parallel: global spatial context aggregation unifies multi-frame features in a shared coordinate system while temporal dynamics extraction models scene evolution via ConvGRU. Their outputs are fused and fed into the decoder, which jointly optimizes historical reconstruction and future forecasting to yield the final 4D occupancy output.

\subsection{Occupancy Encoder}
\label{sec:occ_encoder}
Given $N_h$ historical frames, we denote the occupancy grids and ego-poses as $ \{\mathbf{O}_t\}_{t=T-N_h+1}^{T} $ and $ \{\mathbf{P}_t\}_{t=T-N_h+1}^{T} $, where $ \mathbf{P}_t = (\mathbf{t}_t, \mathbf{q}_t) $ with translation $ \mathbf{t}_t \in \mathbb{R}^3 $ and unit quaternion $ \mathbf{q}_t \in \mathbb{R}^4 $. Following~\cite{zheng2024occworld}, we compress $ \mathbf{O}_t \in \mathbb{R}^{H \times W \times D} $ into a compact BEV representation $ \bB_t \in \mathbb{R}^{c \times h \times w} $, forming $ \{\bB_t\}_{t=T-N_h+1}^{T} $. To model ego-motion, we compute relative motions between consecutive frames~\cite{liao2025i2}. For $ t \in [T - N_h + 1, T - 1] $, the translation delta $ \Delta \mathbf{t}_t \in \mathbb{R}^3 $ and relative rotation $ \Delta \mathbf{q}_t \in \mathbb{R}^4 $ are embedded with other motion information into an ego-motion token via a lightweight pose encoder. This token is fused with BEV features through cross-attention to predict the future $ N_f $ relative pose changes $ \{\Delta \hat{\mathbf{t}}_t, \Delta \hat{\mathbf{q}}_t\}_{t=T+1}^{T+N_f} $, which are integrated to obtain absolute future poses $ \{\hat{\mathbf{P}}_t\}_{t=T+1}^{T+N_f} $. For fair comparison with methods~\cite{liao2025i2, shi2025come, gu2024dome} assuming known future ego-motion, our \shortname can also use ground-truth poses, demonstrating remarkable flexibility.

\subsection{Progressive Explicit-Implicit Generation}
\label{sec:ceig_module}
In this section, we present our progressive explicit-implicit generation (PEIG) module that produces initial per-frame future features by combining explicit geometric propagation with implicit semantic refinement, which preserves geometric fidelity of static structures and captures plausible semantic variations of dynamic scenes.

\noindent\textbf{Explicit Geometric Transformation.}
Given that static scene structures undergo deterministic geometric transformations under ego-motion, we design the explicit geometric transformation (EGT) that leverages ego-poses as strong geometric priors to perform rigid spatial warping on the current BEV features, yielding geometrically consistent initial states for future occupancy. Specifically, given the current BEV feature $ \bB_T $ and the future absolute pose sequence $ \{\hat{\mathbf{P}}_{T+1}, ..., \hat{\mathbf{P}}_{T+N_f}\} $, for each future time step $ t \in [T+1, T+N_f] $, we warp $ \bB_T $ to the future ego-coordinate system via 
\begin{equation}
\label{eq:explicit_warping}
    \bB_t^{geo} = \text{Warp}(\bB_T; \mathbf{P}_T^{-1} \circ \hat{\mathbf{P}}_t),
\end{equation}
where $\circ$ denotes the pose composition operation, and Warp performs differentiable grid sampling. This parameter-free module provides a geometrically faithful static background for subsequent dynamic modeling.

\noindent\textbf{Implicit Feature Modulation.}
To model non-rigid scene dynamics, we propose implicit feature modulation (IFM) that adaptively refines the warped features conditioned on ego-motion priors. Specifically, we leverage the future relative pose changes $ \{\Delta \hat{\mathbf{t}}_t, \Delta \hat{\mathbf{q}}_t\}_{t=T+1}^{T+N_f} $ as prior information for dynamic modeling. For each future time step $ t \in [T+1, T+N_f] $, we concatenate the translation changes $ \Delta \hat{\mathbf{t}}_t \in \mathbb{R}^{3}$ and rotation changes $ \Delta \hat{\mathbf{q}}_t \in \mathbb{R}^{4}$ along the feature dimension, forming complete relative pose change vectors. We encode all temporal relative pose changes through a multi-layer perceptron (MLP), and generate scaling parameters $\gamma_t \in \mmR^{c}$ and offset parameters $\beta_t \in \mmR^{c}$ via two separate linear layers. For each future time step t, we apply a channel-wise affine transformation to the explicitly transformed feature $\bB_t^{geo}$:
\begin{equation}
\label{eq:implicit_modulation}
    \bB_t^{mod} = \gamma_t \odot \bB_t^{geo} + \beta_t,
\end{equation}
where $\odot$ denotes channel-wise multiplication. This module adaptively modulates features using relative pose changes, enhancing dynamic scene representation.

\noindent\textbf{Feature Refinement.}
To refine local spatial details and dynamic scene coherence, we propose a feature refinement (FR) module that employs deformable cross-attention to ground future predictions in high-fidelity context from the current observation. Specifically, given the feature map $F$, each query $q$ will aggregate the sampled features around the corresponding 2D reference point $p$ through 
\begin{equation}
\label{eq:deformattn}
    \text{DeformAttn}(q, p, F) = \sum_{i=1}^{N_{head}} W_i \sum_{j=1}^{N_{point}} A_{ij} W_{i}^{'} F(p +  \triangle p_{ij}),
\end{equation}
where $N_{head}$ and $N_{point}$ denote the number of attention heads and sampled points, respectively. $A_{ij} \in [0,1]$ and $\triangle p_{ij} \in \mathbb R^{2}$ are attention weight and sampling offset for the $j^{th}$ point in the $i^{th}$ head. $F(p + \triangle p_{ij})$ represents the feature of the sampled point $p + \triangle p_{ij}$ obtained by bilinear interpolation. For each future time step $ t \in [T+1, T+N_f] $, we denote the real-world coordinates of the BEV grid centers at t as $\mathbf{C}_t \in \mmR^{h \times w \times 2}$. Leveraging the relative ego pose between t and the current time step T, we apply a coordinate transformation $\mT_{t \to T}$ to project $\mathbf{C}_t$ into the current ego-centric coordinate system, yielding 2D reference points $\mT_{t \to T}(\mathbf{C}_t) \in \mmR^{h \times w \times 2}$. We treat the modulated future BEV feature $\bB_t^{mod}$ as queries, and the current-time BEV feature $\bB_{T}$ as keys and values. The refined features are computed as
\begin{equation}
\label{eq:fea_refine}
    \bB_t^{ref} = \text{DeformAttn}(\bB_t^{mod}, \mT_{t \to T}(\mathbf{C}_t), \bB_{T}).
\end{equation}
The resulting features are then passed through a feed-forward network. This mechanism allows each future BEV location to adaptively attend to semantically and spatially relevant regions in the current scene.

\subsection{Dual-Path Spatio-Temporal Modeling}
\label{sec:dstm_module}
Given per-frame future features, we propose a dual-path spatio-temporal modeling (DPSTM) module for cross-frame modeling via (1) global spatial context aggregation that unifies multi-frame features in a shared coordinate system for geometric consistency, and (2) temporal dynamics extraction that models semantic evolution through sequential processing for temporal coherence.

\noindent\textbf{Global Spatial Context Aggregation.}  
To capture long-range spatial dependencies and enrich scene-level semantic understanding, we design the global spatial context aggregation (GSCA) that leverages both historical and future ego-poses to unify multi-frame BEV features into a shared global coordinate system. Specifically, we transform the historical features $ \{\bB_t\}_{t=T-N_h+1}^{T} $ and the refined future features $ \{\bB_t^{ref}\}_{t=T+1}^{T+N_f} $ from their ego-centric frames into a fixed global coordinate system using their corresponding poses $ \{\mathbf{P}_t\}_{t=T-N_h+1}^{T} $ and $ \{\hat{\mathbf{P}}_t\}_{t=T+1}^{T+N_f} $, yielding a temporally extended sequence of spatially aligned feature maps $ \{\bB_t^{align}\}_{t=T-N_h+1}^{T+N_f} $. To capture multi-granularity context, we project the aligned feature maps onto BEV grids at $K$ decreasing resolutions $\{r_k\}_{k=1}^{K}$, where $r_1$ is the coarsest and $r_k$ the finest, forming a multi-scale BEV representation. At each scale $k$, a lightweight convolutional block aggregates local spatial information, producing scale-specific global features $\bF_k^{global}$. Low-resolution features ($k=1$) encode large-scale scene topology, while high-resolution features ($k=K$) preserve fine-grained details. We then fuse these features hierarchically in a coarse-to-fine manner. Starting with $\tilde{\bF}_1 = \bF_1^{global}$, we iteratively upsample, concatenate, and refine via a residual convolution block:
\begin{equation}
\label{eq:fea_upsample}
    \tilde{\bF}_{k} = \text{Conv}(\text{Upsample}(\tilde{\bF}_{k-1}) \oplus \bF_k^{global}), \quad k = 2, \dots, K,
\end{equation}
where $\oplus$ denotes channel-wise concatenation. The final full-resolution global contextual feature is defined as $\bF^{global} := \tilde{\bF}_K$. We map it back to the ego-centric BEV grid of each future time step $t \in [T+1, T+N_f]$ via differentiable grid sampling:
\begin{equation}
\label{eq:fea_gridsample}
    \bB_t^{spatial} = \text{GridSample}(\bF^{global},\; \phi(\hat{\mathbf{P}}_t)),
\end{equation}
where $\phi(\hat{\mathbf{P}}_t)$ denotes the sampling grid obtained by transforming the ego-centric BEV coordinates at time t into the global frame using the predicted pose $\hat{\mathbf{P}}_t$, followed by normalization to $[-1,1]$. This process yields globally consistent, context-aware future BEV features $\{\bB_t^{spatial}\}_{t=T+1}^{T+N_f}$.

\noindent\textbf{Temporal Dynamics Extraction.}  
To model the dynamic evolution of scene semantics, we introduce the temporal dynamics extraction (TDE) that propagates contextual information across historical and future frames. Specifically, we concatenate the historical BEV features $ \{\bB_t\}_{t=T-N_h+1}^{T} $ and the refined future features $ \{\bB_t^{ref}\}_{t=T+1}^{T+N_f} $ along the time dimension to form a continuous spatio-temporal sequence $\mathcal{B}_{T-N_h+1:T+N_f} = [ \bB_{T-N_h+1}, \dots, \bB_T, \bB_{T+1}^{ref}, \dots, \bB_{T+N_f}^{ref} ]$. We then process $\mathcal{B}$ in chronological order using a lightweight ConvGRU encoder. Denoting the hidden state at time step $t$ as $h_t$, the recurrence is defined as:
\begin{equation}
\label{eq:fea_gru}
    \mathbf{h}_t = \text{ConvGRU}(\mathcal{B}_t, \mathbf{h}_{t-1}), \quad t = T - N_h + 1, \dots, T + N_f,
\end{equation}
with $\mathbf{h}_{T-N_h} = \mathbf{0}$. The hidden state $\mathbf{h}_t$ serves as a temporally aware representation, encoding the contextual evolution from the start of the sequence up to time $t$. For future steps ($t > T$), it propagates historical dynamics forward in a causal manner. Finally, we project each hidden state to the BEV feature space to obtain dynamic-aware representations $\{\bB_t^{temp}\}_{t=T+1}^{T+N_f}$. This module enhances the temporal coherence and dynamic plausibility of future predictions.

Global spatial context aggregation and temporal dynamics extraction serve as complementary branches: the former enforces cross-frame geometric consistency while the latter ensures temporal smoothness. Their outputs are fused at each future time step $t \in [T+1, T+N_f]$ via channel-wise concatenation followed by a convolutional block to produce the final future BEV features:
\begin{equation}
\label{eq:fea_concat}
    \hat{\bB}_t = \text{Conv}(\bB_t^{spatial} \,\oplus\, \bB_t^{temp}),
\end{equation}
where $\oplus$ denotes channel-wise concatenation. We further enhance future BEV features with deformable self-attention to exploit intra-frame semantics.

\subsection{Occupancy Decoder}
\label{sec:occ_decoder}
To facilitate spatio-temporal representation learning, our decoder jointly optimizes two objectives: reconstructing historical occupancy from past BEV features and predicting future occupancy from enhanced future features. Specifically, we construct a unified spatio-temporal feature volume by concatenating historical BEV features $\{\bB_t\}_{t=T-N_h+1}^{T}$ and enhanced future features $\{\hat{\bB}_t\}_{t=T+1}^{T+N_f}$ along the time dimension. This volume is upsampled to the original resolution via bilinear interpolation and subsequently processed by residual convolution blocks to predict 4D occupancy $\hat{\bO}_{T-N_h+1:T+N_f} \in \mmR^{(N_h+N_f) \times S \times H \times W \times D}$, with $S$ semantic classes. The historical segment ($t \leq T$) serves as a reconstruction target, while the future segment ($t > T$) provides the final predictions. This design encourages consistent, high-fidelity representations across both observed and unobserved time steps, bridging perception and prediction in a unified framework.

\subsection{Optimization and Training Details}
We train our \shortname in an end-to-end manner with an objective that combines occupancy prediction loss and ego-pose regression loss. For 4D occupancy, we supervise both historical reconstruction (\(t \leq T\)) and future forecasting (\(t > T\)) against ground-truth labels $\bO_{T-N_h+1:T+N_f} \in \mmR^{(N_h+N_f) \times H \times W \times D}$, using a weighted cross-entropy loss and a Lovász-Softmax loss for each segment. The total occupancy loss is the sum of the historical and future components, each weighted by the same hyperparameters \(\lambda_{\text{wce}}\) and \(\lambda_{\text{lov}}\). For ego-motion, the encoder predicts future relative pose changes (translation deltas and rotation quaternions). The pose loss combines an L2 loss for translation and a quaternion-based angular loss, with weights \(\lambda_{\text{trans}} = 0.01\) and \(\lambda_{\text{rot}} = 1.0\) following~\cite{liao2025i2}. The overall objective is the sum of the occupancy and pose losses.
\section{Experiments}
\label{sec:experiments}

\subsection{Experimental Setup}

\noindent\textbf{Dataset.}
Following~\cite{zheng2024occworld,gu2024dome,shi2025come,liao2025i2}, we evaluate our method on the widely used Occ3D-nuScenes~\cite{tian2023occ3d}, a semantic occupancy prediction benchmark derived from the large-scale nuScenes~\cite{caesar2020nuscenes}. Occ3D-nuScenes provides semantic annotations across 18 categories, comprising 17 semantic classes and 1 empty class. The spatial coverage of the dataset ranges from $[-40m, -40m, -1m]$ to $[40m, 40m, 5.4m]$ along the X, Y, and Z axes, respectively. Given a voxel size of $0.4m\times0.4m\times0.4m$, this volume is discretized into a $200\times200\times16$ voxel grid. This dataset consists of 700 training scenes and 150 validation scenes. Moreover, we perform zero-shot evaluation on the Occ3D-Waymo~\cite{tian2023occ3d} validation set derived from the Waymo Open Dataset~\cite{sun2020scalability}, which comprises 202 validation scenes and has the same spatial range and voxel resolution as Occ3D-nuScenes.

\noindent\textbf{Evaluation metrics.}
We adopt the intersection over union (IoU) as the geometric metric, distinguishing between occupied and empty voxels. Meanwhile, we report the mean intersection over union (mIoU) across all semantic classes to evaluate the quality of semantic segmentation on occupied voxels. Following the evaluation protocol of~\cite{zheng2024occworld, liao2025i2}, we present results for each future timestamp and the average performance across all timestamps on the validation set.

\noindent\textbf{Implementation details.}
We implement our method using PyTorch~\cite{paszke2019pytorch}. Following prior work~\cite{zheng2024occworld,gu2024dome,shi2025come,liao2025i2}, we leverage 2 seconds of historical data ($N_h=4$) to predict the future occupancy for the next 3 seconds ($N_f=6$). We employ one deformable cross-attention layer (with 4 attention heads and 8 sampled points) for feature refinement and two ConvGRU layers for temporal dynamics extraction. The number of BEV resolutions $K$ for global spatial context aggregation is set to 3, corresponding to spatial grids of $50\times50$, $100\times100$, and $200\times200$. The feature dimension is set to 128 empirically. We train the model for 30 epochs using the AdamW optimizer~\cite{loshchilov2017decoupled} with a weight decay of 0.01. The initial learning rate is set to 0.001 and decays to zero following a cosine annealing schedule. All experiments are conducted with a batch size of 8 across 4 NVIDIA RTX 3090 GPUs. The loss weighting coefficients $\lambda_{wce}$ and $\lambda_{lov}$ are all set to 1.0. 

\begin{table*}[ht]
    \caption{4D occupancy forecasting performance on Occ3D-nuScenes validation set. The settings vary by input modality and whether the ego trajectory is predicted (Pred.) or provided as ground truth (GT). ``Avg." denotes the average performance across 1s, 2s, and 3s horizons. The \textbf{bold} numbers indicate the best results.}
    \centering
    \scalebox{0.90}{
    \begin{tabular}{l|cc|ccc>{\columncolor[gray]{0.92}}c|ccc>{\columncolor[gray]{0.92}}c}
        \hline
        \multirow{2}{*}{Method} & \multirow{2}{*}{Input} & \multirow{2}{*}{Ego traj.} &
        \multicolumn{4}{c|}{mIoU (\%) $\uparrow$} & 
        \multicolumn{4}{c}{IoU (\%) $\uparrow$} \\
        & & & 1s & 2s & 3s & Avg. & 1s & 2s & 3s & Avg. \\
        \hline\hline
        OccWorld-D~\cite{zheng2024occworld} & Camera & Pred. & 11.55 & 8.10 & 6.22 & 8.62 & 18.90 & 16.26 & 14.43 & 16.53 \\
        OccWorld-T~\cite{zheng2024occworld} & Camera & Pred. & 4.68 & 3.36 & 2.63 & 3.56 & 9.32 & 8.23 & 7.47 & 8.34 \\
        OccWorld-S~\cite{zheng2024occworld} & Camera & Pred. & 0.28 & 0.26 & 0.24 & 0.26 & 5.05 & 5.01 & 4.95 & 5.00 \\
        OccWorld-F~\cite{zheng2024occworld} & Camera & Pred. & 8.03 & 6.91 & 3.54 & 6.16 & 23.62 & 18.13 & 15.22 & 18.99 \\
        RenderWorld~\cite{yan2025renderworld} & Camera & Pred. & 2.83 & 2.55 & 2.37 & 2.58 & 14.61 & 13.61 & 12.98 & 13.73 \\
        OccLLaMA~\cite{wei2024occllama} & Camera & Pred. & 10.34 & 8.66 & 6.98 & 8.66 & 25.81 & 23.19 & 19.97 & 22.99 \\
        PreWorld~\cite{li2025semi} & Camera & Pred. & 12.27 & 9.24 & 7.15 & 9.55 & 23.62 & 21.62 & 19.63 & 21.62 \\
        Occ-LLM~\cite{xu2025occ} & Camera & Pred. & 11.28 & 10.21 & 9.13 & 10.21 & 27.11 & 24.07 & 20.19 & 23.79 \\
        DFIT-OccWorld~\cite{zhang2024efficient} & Camera & Pred. & 13.38 & 10.16 & 7.96 & 10.50 & 19.18 & 16.85 & 15.02 & 17.02 \\
        \shortname-STC (Ours) & Camera & Pred. & \textbf{19.32} & \textbf{13.96} & \textbf{10.62} & \textbf{14.84} & \textbf{27.97} & \textbf{23.35} & \textbf{20.22} & \textbf{23.87} \\

        \hline
        DOME-STC~\cite{gu2024dome} & Camera & GT & 17.79 & 14.23 & 11.58 & 14.53 & 26.39 & 23.20 & 20.42 & 23.33 \\
        $I^{2}$-World-STC~\cite{liao2025i2} & Camera & GT & 21.67 & 18.78 & 16.47 & 18.97 & 30.55 & 28.76 & 26.99 & 28.77 \\
        \shortname-STC (Ours) & Camera & GT & \textbf{22.90} & \textbf{19.93} & \textbf{17.17} & \textbf{20.16} & \textbf{31.62} & \textbf{29.92} & \textbf{28.00} & \textbf{29.87} \\

        \hline
        Copy\&Paste~\cite{zheng2024occworld} & 3D-Occ & Pred. & 14.91 & 10.54 & 8.52 & 11.33 & 24.47 & 19.77 & 17.31 & 20.52 \\ 
        OccWorld~\cite{zheng2024occworld} & 3D-Occ & Pred. & 25.78 & 15.14 & 10.51 & 17.14 & 34.63 & 25.07 & 20.18 & 26.63 \\
        OccLLaMA~\cite{wei2024occllama} & 3D-Occ & Pred. & 25.05 & 19.49 & 15.26 & 19.93 & 34.56 & 28.53 & 24.41 & 29.17 \\
        RenderWorld~\cite{yan2025renderworld} & 3D-Occ & Pred. & 28.69 & 18.89 & 14.83 & 20.80 & 37.74 & 28.41 & 24.08 & 30.08 \\
        Occ-LLM~\cite{xu2025occ} & 3D-Occ & Pred. & 24.02 & 21.65 & 17.29 & 20.99 & 36.65 & 32.14 & 28.77 & 32.52 \\
        COME~\cite{shi2025come} & 3D-Occ & Pred. & 30.57 & 19.91 & 13.38 & 21.29 & 36.96 & 28.26 & 21.86 & 29.03 \\
        DFIT-OccWorld~\cite{zhang2024efficient} & 3D-Occ & Pred. & 31.68 & 21.29 & 15.18 & 22.71 & 40.28 & 31.24 & 25.29 & 32.27 \\
        \shortname (Ours) & 3D-Occ & Pred. & \textbf{38.60} & \textbf{24.79} & \textbf{18.54} & \textbf{27.38} & \textbf{44.69} & \textbf{34.21} & \textbf{28.85} & \textbf{35.90} \\
        
        \hline
        DOME~\cite{gu2024dome} & 3D-Occ & GT & 35.11 & 25.89 & 20.29 & 27.10 & 43.99 & 35.36 & 29.74 & 36.36 \\
        UniScene~\cite{li2025uniscene} & 3D-Occ & GT & 35.37 & 29.59 & 25.08 & 31.76 & 38.34 & 32.70 & 29.09 & 34.84 \\
        COME~\cite{shi2025come} & 3D-Occ & GT & 42.75 & 32.97 & 26.98 & 34.23 & 50.57 & 43.47 & 38.36 & 44.13 \\
        $I^{2}$-World~\cite{liao2025i2} & 3D-Occ & GT & 47.62 & 38.58 & 32.98 & 39.73 & 54.29 & 49.43 & 45.69 & 49.80 \\
        \shortname (Ours) & 3D-Occ & GT & \textbf{55.54} & \textbf{46.33} & \textbf{40.18} & \textbf{47.40} & \textbf{60.77} & \textbf{55.97} & \textbf{51.96} & \textbf{56.24} \\
    
        \hline
    \end{tabular}
    }
\label{tab:occ3d_nuscenes}
\end{table*}

\begin{table*}[ht]
    \caption{Long-term 4D occupancy forecasting performance on Occ3D-nuScenes validation set. Ground-truth 3D occupancy and trajectories are used as inputs. ``Avg.” denotes the average performance over the 1-8 second prediction horizon. The \textbf{bold} numbers indicate the best results.}
    \centering
    \scalebox{0.90}{
    \begin{tabular}{l|cccccccc|>{\columncolor[gray]{0.92}}c}
        \hline
        \multirow{2}{*}{Method} & 
        \multicolumn{9}{c}{mIoU (\%) $\uparrow$} \\
        & 1s & 2s & 3s & 4s & 5s & 6s & 7s & 8s & Avg. \\
        \hline
        DOME~\cite{gu2024dome} & 30.10 & 21.35 & 17.36 & 14.86 & 12.61 & 11.03 & 10.00 & 9.34 & 15.83 \\
        COME~\cite{shi2025come} & 33.78 & 24.57 & 21.35 & 18.25 & 15.84 & 13.85 & 12.99 & 11.96 & 19.07 \\
        \shortname (Ours) & \textbf{36.23} & \textbf{30.61} & \textbf{27.13} & \textbf{24.36} & \textbf{21.98} & \textbf{19.81} & \textbf{17.83} & \textbf{15.96} & \textbf{24.12} \\
        
        \hline\hline
        \multirow{2}{*}{Method} & 
        \multicolumn{9}{c}{IoU (\%) $\uparrow$} \\
        & 1s & 2s & 3s & 4s & 5s & 6s & 7s & 8s & Avg. \\
        \hline
        DOME~\cite{gu2024dome} & 39.04 & 31.20 & 27.14 & 24.73 & 22.32 & 20.28 & 19.05 & 17.97 & 25.21 \\
        COME~\cite{shi2025come} & 44.20 & 36.25 & 32.86 & 30.03 & 26.93 & 24.70 & 23.30 & 21.44 & 29.96 \\
        \shortname (Ours) & \textbf{49.09} & \textbf{45.11} & \textbf{42.11} & \textbf{39.52} & \textbf{37.19} & \textbf{34.99} & \textbf{32.87} & \textbf{30.87} & \textbf{39.06} \\
    
        \hline
    \end{tabular}
    }
\label{tab:long_term}
\end{table*}

\subsection{Quantitative Comparison}

\noindent\textbf{4D occupancy forecasting.}
We evaluate our method on the Occ3D-nuScenes validation set. Following~\cite{zheng2024occworld,gu2024dome,shi2025come,liao2025i2}, we assess our \shortname under various configurations, determined by two key factors: (1) whether the model input consists of ground-truth 3D occupancy or raw sensor inputs (\eg, camera images), and (2) whether the ego trajectory is ground truth or generated by a planning module. The results are presented in Table~\ref{tab:occ3d_nuscenes}. When camera images are used as model input, we follow the established approach of $I^{2}$-World~\cite{liao2025i2} by employing STCOcc~\cite{liao2025stcocc} to process the visual data and generate semantic occupancy predictions. For scenarios where the ego trajectory is predicted, we forecast the future ego poses by leveraging historical scene features and ego-motion information. Under this configuration, our \shortname achieves the best performance, with an average mIoU of 14.84\% and an average IoU of 23.87\%. For example, it surpasses DFIT-OccWorld~\cite{zhang2024efficient} by 4.34\% in mIoU and by 6.85\% in IoU. When the ground-truth ego trajectory is provided, we perform a comparative evaluation with DOME-STC~\cite{gu2024dome} and $I^{2}$-World-STC~\cite{liao2025i2}, with all methods using predictions from STCOcc~\cite{liao2025stcocc}. Specifically, across the 1s, 2s, and 3s prediction horizons, our method consistently delivers the highest accuracy, yielding mIoU of 22.90\%, 19.93\%, 17.17\%, and IoU of 31.62\%, 29.92\%, 28.00\%, respectively. When the model is provided with ground-truth 3D occupancy along with a predicted ego trajectory, our \shortname achieves an average mIoU of 27.38\% and an average IoU of 35.90\%, substantially outperforming all existing approaches. Further improvements are observed when ground-truth 3D occupancy and ground-truth ego trajectory are used simultaneously, elevating performance to 47.40\% mIoU and 56.24\% IoU, corresponding to gains of 20.02\% mIoU and 20.34\% IoU, respectively. Compared to state-of-the-art methods, including diffusion-based approaches~\cite{gu2024dome,shi2025come,li2025uniscene} and autoregressive frameworks~\cite{liao2025i2}, our \shortname demonstrates consistently stronger predictive accuracy. Specifically, it surpasses $I^{2}$-World~\cite{liao2025i2} by a notable margin of 7.67\% in mIoU and 6.44\% in IoU. These consistent and significant improvements across diverse evaluation settings highlight not only the robustness of our design under practical conditions but also its superior capacity to capture spatio-temporal dependencies in dynamic 3D scenes, thereby validating the effectiveness and advancement of the proposed method.

\noindent\textbf{Long-term forecasting.}
Long-term occupancy forecasting plays a crucial role in autonomous driving, enabling safer planning and decision-making in dynamic environments. To evaluate the long-term predictive capability of our method, we extend the prediction horizon from 3 seconds to 8 seconds and use both ground-truth 3D occupancy and ground-truth ego trajectory as model input, following the setup of COME~\cite{shi2025come}. As shown in Table~\ref{tab:long_term}, our \shortname achieves an average mIoU of 24.12\% and an average IoU of 39.06\% over the full 8-second horizon, consistently surpassing all compared baselines at every future timestamp. Specifically, it outperforms DOME~\cite{gu2024dome} and COME~\cite{shi2025come} by 8.29\%, 5.05\% in mIoU and 13.85\%, 9.1\% in IoU, respectively. These results thus validate the effectiveness and scalability of our method in long-term forecasting.

\begin{table}[ht]
\centering
\begin{minipage}{0.38\textwidth}
    \centering
    \caption{Zero-shot performance on Occ3D-Waymo validation set.}
    \scalebox{0.76}{
        \begin{tabular}{c|c|cc}
        \hline
        Method & Rate & mIoU (\%) & IoU (\%) \\
        \hline
        Copy\&Paste & 10Hz & 28.34 & 40.09 \\
        $I^{2}$-World~\cite{liao2025i2} & 10Hz & 43.73 & 60.97 \\
        \shortname (Ours) & 10Hz & \textbf{57.47} & \textbf{70.66} \\
        \hline
        Copy\&Paste & 2Hz & 17.17 & 28.21 \\
        $I^{2}$-World~\cite{liao2025i2} & 2Hz & 36.38 & 52.36 \\
        \shortname (Ours) & 2Hz & \textbf{46.70} & \textbf{60.09} \\
        \hline
        \end{tabular}
    }
    \label{tab:occ3d_waymo}
\end{minipage}
\hfill
\begin{minipage}{0.60\textwidth}
    \centering
    \caption{Comparisons of model efficiency on the Occ3D-nuScenes validation set. The \textbf{bold} numbers indicate the best results.}
    \scalebox{0.76}{
        \begin{tabular}{c|ccccc}
        \hline
        Method & \#Params.$\downarrow$ & \#FLOPs$\downarrow$ & Latency$\downarrow$ & mIoU$\uparrow$ & IoU$\uparrow$ \\
        \hline
        DOME~\cite{gu2024dome} & 444.07M & 2928.66G & 1899.15ms & 27.10\% & 36.36\% \\
        COME~\cite{shi2025come} & 692.97M & 4461.46G & 4377.67ms & 34.23\% & 44.13\% \\
        $I^{2}$-World~\cite{liao2025i2} & 22.67M & 494.58G & 227.09ms & 39.73\% & 49.80\% \\
        \hline
        \shortname (Ours) & \textbf{12.09M} & \textbf{416.27G} & \textbf{80.03ms} & \textbf{47.40\%} & \textbf{56.24\%} \\
        \hline
        \end{tabular}
    }
    \label{tab:efficiency_analysis}
\end{minipage}
\end{table}

\begin{figure*}[ht]
    \centering
    \includegraphics[width=1.0\linewidth]{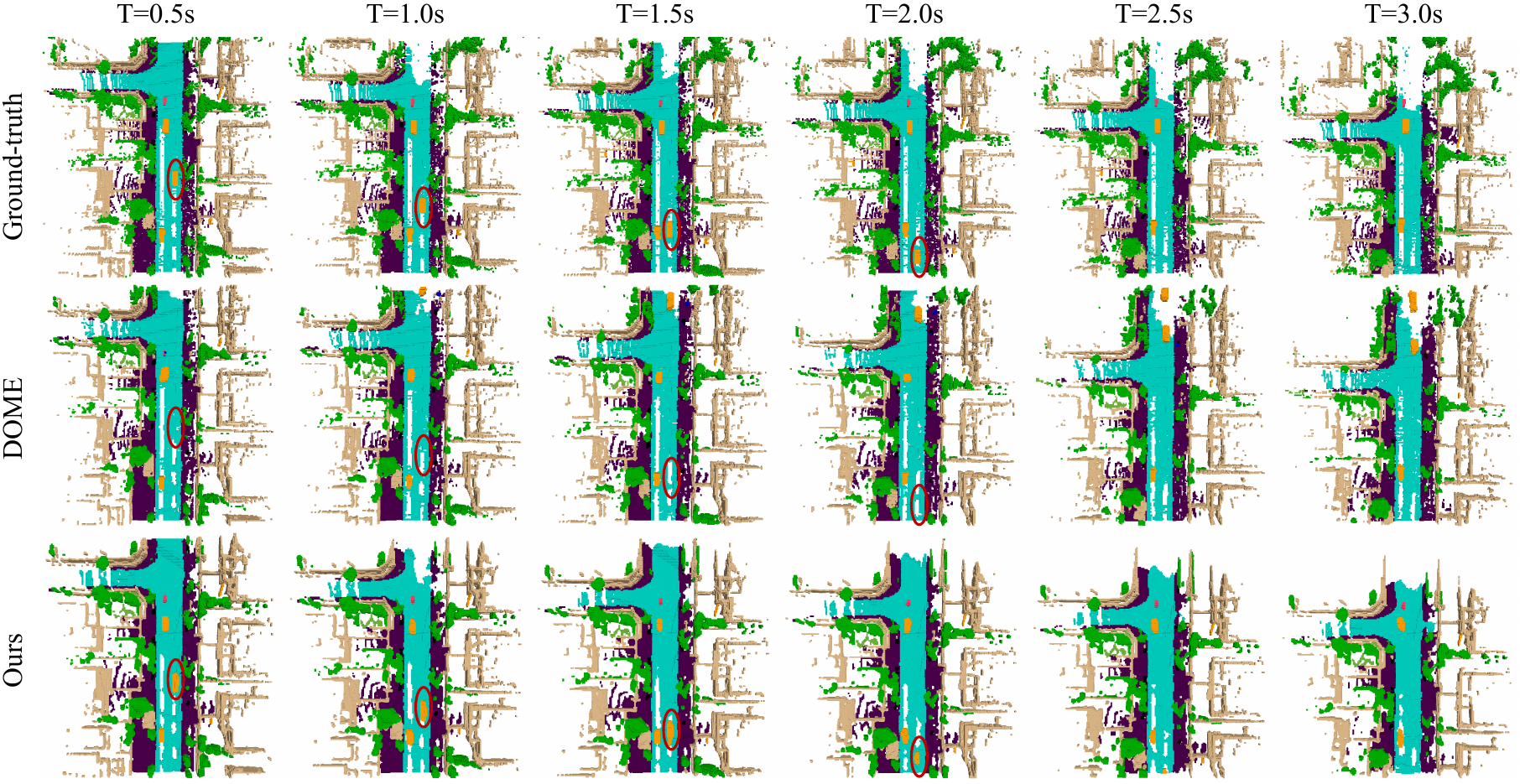}
    \vspace{-0.6cm}
    \caption{Qualitative results of \shortname on Occ3D-nuScenes.}
    \label{fig:nus_vis_cmp}
\end{figure*}

\noindent\textbf{Zero-shot forecasting.}
To evaluate the generalization ability of our method, we conduct zero-shot evaluation on the Occ3D-Waymo validation set following $I^{2}$-World~\cite{liao2025i2}. As shown in Table~\ref{tab:occ3d_waymo}, our \shortname delivers superior performance under different sampling rates. Specifically, at the 10 Hz setting, our \shortname surpasses $I^{2}$-World~\cite{liao2025i2} by 13.74\% in mIoU and 9.69\% in IoU. When the sampling rate drops to 2 Hz, the margins are 10.32\% and 7.73\% in mIoU and IoU, respectively. These improvements demonstrate the strong generalization of our method.

\subsection{Qualitative Evaluation}
To better understand the benefits of our method, we provide qualitative evaluations on Occ3D-nuScenes. As shown in Figure~\ref{fig:nus_vis_cmp}, our \shortname achieves superior geometric fidelity and temporal consistency, particularly in preserving fine-grained structural details and capturing dynamic object motions, which demonstrates its effectiveness in modeling both static structure and dynamic motion.

\begin{table}[ht]
\centering
\begin{minipage}{0.56\textwidth}
    \centering
    \caption{Ablation study of the proposed model components on Occ3D-nuScenes validation set. 
    The \textbf{bold} numbers denote the best results.}
    \scalebox{0.94}{
        \begin{tabular}{cccc|cc|cc}
        \hline
        & \multicolumn{3}{c|}{PEIG} & \multicolumn{2}{c|}{DPSTM} & \multirow{2}{*}{mIoU (\%)} & \multirow{2}{*}{IoU (\%)} \\
        & EGT & IFM & FR & GSCA & TDE & & \\
        \hline
        1 & & & & & & 18.31 & 29.01 \\
        2 & $\checkmark$ & & & & & 31.78 & 41.65 \\
        3 & $\checkmark$ & $\checkmark$ & & & & 32.55 & 42.63 \\
        4 & $\checkmark$ & $\checkmark$ & $\checkmark$ & & & 40.47 & 49.15 \\
        5 & $\checkmark$ & $\checkmark$ & $\checkmark$ & $\checkmark$ & & 46.08 & 55.50 \\
        6 & $\checkmark$ & $\checkmark$ & $\checkmark$ & $\checkmark$ & $\checkmark$ & \textbf{47.40} & \textbf{56.24} \\
        \hline
        \end{tabular}
        }
    \label{tab:model_components}
\end{minipage}
\hfill
\begin{minipage}{0.40\textwidth}
    \centering
    \caption{Effect of the number of historical frames on the Occ3D-nuScenes validation set.}
    \scalebox{0.84}{
        \begin{tabular}{c|c|cc}
        \hline
        Number & Ego traj. & mIoU (\%) & IoU (\%) \\
        \hline
        1 & Pred. & 17.01 & 30.46 \\
        2 & Pred. & 24.51 & 35.11 \\
        3 & Pred. & 26.32 & 35.50 \\
        4 & Pred. & \textbf{27.38} & \textbf{35.90} \\
        \hline
        1 & GT & 34.88 & 51.52 \\
        2 & GT & 46.18 & 55.91 \\
        3 & GT & 46.97 & 55.85 \\
        4 & GT & \textbf{47.40} & \textbf{56.24} \\
        \hline
        \end{tabular}
        }
    \label{tab:num_hisframe}
\end{minipage}
\end{table}

\subsection{Efficiency Analysis}
We evaluate the efficiency of our method on the Occ3D-nuScenes validation set using a single GeForce RTX 3090. As shown in Table~\ref{tab:efficiency_analysis}, our \shortname achieves better performance with fewer model parameters, lower FLOPs, and reduced inference latency, demonstrating its effectiveness and efficiency. For instance, our \shortname is $2.84\times$ faster than $I^{2}$-World~\cite{liao2025i2} while delivering a 7.67\% mIoU improvement.

\subsection{Ablation Study}

\noindent\textbf{Effect of the proposed model components.}
We study the effect of the proposed model components of our \shortname on the Occ3D-nuScenes validation set, including explicit geometric transformation (EGT), implicit feature modulation (IFM), feature refinement (FR) of progressive explicit-implicit generation (PEIG), and global spatial context aggregation (GSCA), temporal dynamics extraction (TDE) of dual-path spatio-temporal modeling (DPSTM). Note that the baseline model uses independent per-frame convolutions to predict future occupancy from current BEV features. The experimental results are shown in Table~\ref{tab:model_components}. Comparing the first and second rows, introducing explicit geometric transformation significantly improves the mIoU by 13.47\% (31.78\% vs. 18.31\%) and IoU by 12.64\% (41.65\% vs. 29.01\%), confirming the importance of geometric priors in preserving static structure consistency. Further integrating implicit feature modulation yields additional gains (32.55\% mIoU, 42.63\% IoU), indicating the benefit of motion-aware feature adaptation. The subsequent feature refinement substantially boosts model performance to 40.47\% mIoU and 49.15\% IoU, which validates its role in fusing contextual cues for local geometry enhancement. Crucially, the progressive explicit-implicit generation module, through the synergy of rigid propagation and elastic modulation, achieves clear improvements over the baseline in per-frame occupancy forecasting. Moreover, from the fourth and fifth rows, the proposed global spatial context aggregation further contributes 5.61\% mIoU (46.08\% vs. 40.47\%) and 6.35\% IoU (55.50\% vs. 49.15\%), highlighting the significance of modeling long-range spatial dependencies in a unified world coordinate system for coherent scene reconstruction. Finally, when combined with temporal dynamics extraction, the full model achieves the best performance at 47.40\% mIoU and 56.24\% IoU, verifying the effectiveness of progressive explicit-implicit generation and dual-path spatio-temporal modeling in producing high-fidelity and temporally consistent 4D occupancy prediction.

\begin{table}[ht]
\centering
\begin{minipage}{0.47\textwidth}
    \centering
    \caption{Effect of the number of DCA layers in feature refinement on Occ3D-nuScenes validation set.}
    \scalebox{1.0}{
        \begin{tabular}{c|ccccc}
        \hline
        Number & 1 & 2 & 3 & 4 \\
        \hline
        mIoU (\%) & 47.40 & 47.62 & \textbf{47.71} & 47.55 \\
        IoU (\%) & 56.24 & \textbf{56.98} & 56.88 & 56.68 \\
        \hline
        \end{tabular}
        }
    \label{tab:num_dca}
\end{minipage}
\hfill
\begin{minipage}{0.49\textwidth}
    \centering
    \caption{Effect of the number of ConvGRU layers in temporal dynamics extraction on Occ3D-nuScenes validation set.}
    \scalebox{1.0}{
        \begin{tabular}{c|ccccc}
        \hline
        Number & 1 & 2 & 3 & 4 \\
        \hline
        mIoU (\%) & 47.03 & 47.40 & 47.31 & \textbf{47.45} \\
        IoU (\%) & 56.04 & 56.24 & 56.74 & \textbf{56.85} \\
        \hline
        \end{tabular}
        }
    \label{tab:num_gru}
\end{minipage}
\end{table}

\noindent\textbf{Effect of the number of historical frames.}
We investigate the impact of the number of historical frames on the Occ3D-nuScenes validation set. As shown in Table~\ref{tab:num_hisframe}, increasing the number of historical frames consistently improves prediction accuracy by providing a richer temporal context for modeling scene dynamics. Specifically, when using the predicted trajectory, mIoU rises from 17.01\% with one frame to 27.38\% with four frames. When the ground-truth trajectory is provided, the model performance further increases to 47.40\% mIoU with four frames. These results demonstrate that expanding the historical horizon enables the model to better capture scene evolution and motion patterns, leading to more accurate future occupancy prediction.

\noindent\textbf{Effect of the number of DCA layers in feature refinement.}
We investigate the impact of the number of deformable cross-attention layers in feature refinement on the Occ3D-nuScenes validation set. As shown in Table~\ref{tab:num_dca}, deformable cross-attention is essential for adaptively aggregating high-fidelity contextual cues from the current observation, thereby enhancing both the local geometry and semantics of the predicted occupancy. Although stacking more layers yields consistent gains in both mIoU and IoU, we adopt a single layer for feature refinement to balance model performance with computational efficiency.

\noindent\textbf{Effect of the number of ConvGRU layers in temporal dynamics extraction.}
We investigate the impact of the number of ConvGRU layers in temporal dynamics extraction on the Occ3D-nuScenes validation set. As shown in Table~\ref{tab:num_gru}, the temporal dynamics extraction module, built upon ConvGRU layers, is crucial for capturing scene evolution by modeling sequential dependencies across frames. Performance improves progressively as the number of ConvGRU layers increases from one to four, with four layers achieving the peak results (47.45\% mIoU, 56.85\% IoU). Considering the gains beyond two layers are marginally incremental, we use two ConvGRU layers as an optimal trade-off between temporal modeling capacity and inference efficiency.

\noindent\textbf{Effect of BEV resolution in global spatial context aggregation.}
We study the effect of BEV resolution in global spatial context aggregation on the Occ3D-nuScenes validation set. This module adopts a multi-scale design with three hierarchical BEV resolutions $50\times50$, $100\times100$, and $200\times200$. As shown in Table~\ref{tab:bev_sizes}, the multi-scale BEV aggregation achieves the overall best performance (47.40\% mIoU, 56.24\% IoU), consistently outperforming all single-resolution variants. The results demonstrate that combining features from coarse-level scene layout and fine-grained structural details enables more comprehensive spatial context modeling, thereby improving occupancy prediction accuracy. 

\begin{table}[ht]
\centering
\begin{minipage}{0.50\textwidth}
    \centering
    \caption{Effect of BEV resolution in global spatial context aggregation on the Occ3D-nuScenes validation set.}
    \scalebox{0.90}{
        \begin{tabular}{ccc|cc}
        \hline
        \multicolumn{3}{c|}{BEV Resolution} & \multirow{2}{*}{mIoU (\%)} & \multirow{2}{*}{IoU (\%)} \\
        50$\times$50 & 100$\times$100 & 200$\times$200 & & \\
        \hline
        $\checkmark$ & & & 46.90 & 55.63 \\
        & $\checkmark$ & & 47.18 & 55,64 \\
        & & $\checkmark$ & 47.17 & 55.73 \\
        \hline
        $\checkmark$ & $\checkmark$ & $\checkmark$ & \textbf{47.40} & \textbf{56.24} \\
        \hline
        \end{tabular}
    }
    \label{tab:bev_sizes}
\end{minipage}
\hfill
\begin{minipage}{0.46\textwidth}
    \centering
    \includegraphics[width=0.88\linewidth]{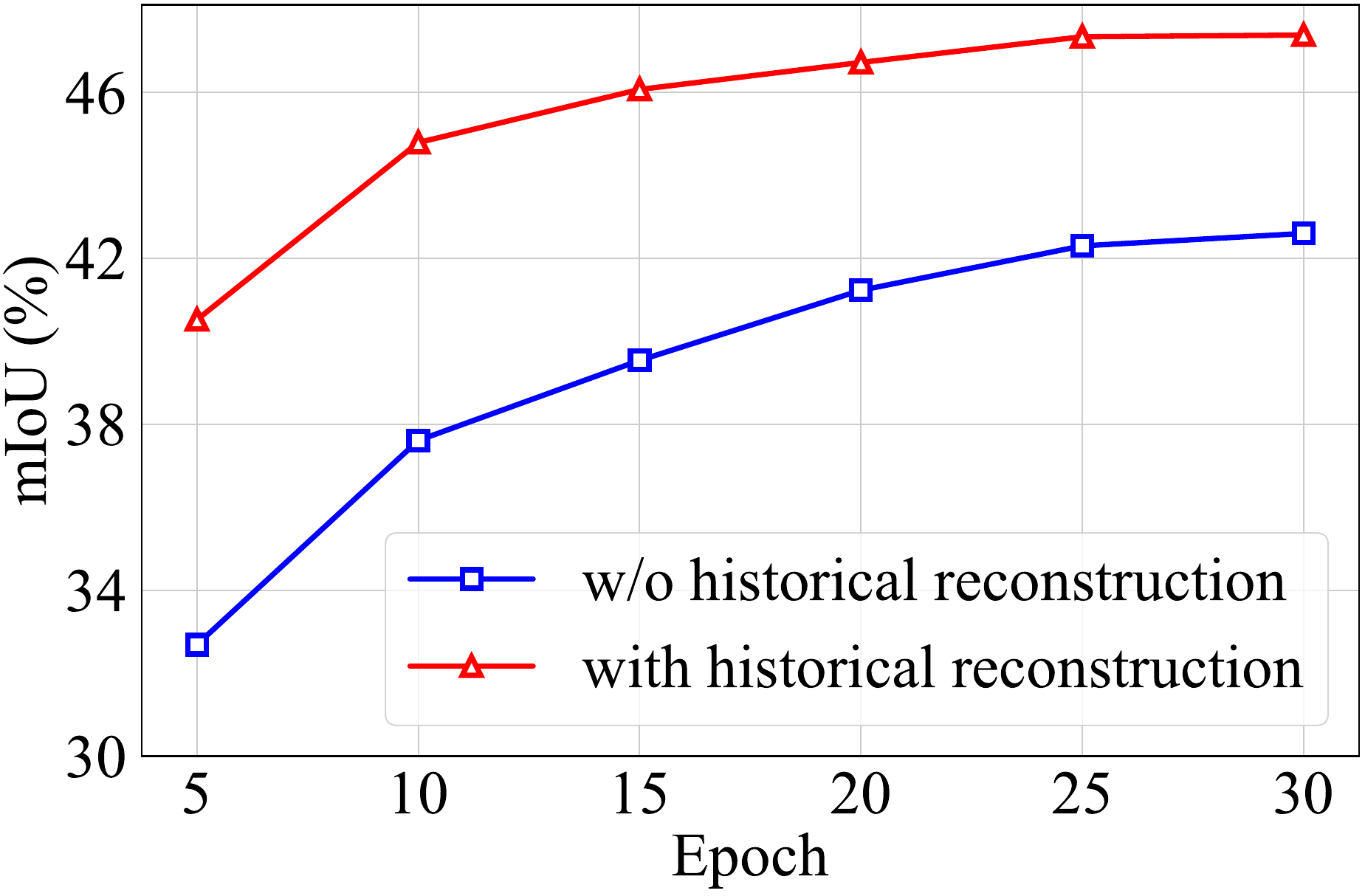}
    \captionof{figure}{Validation curves of mIoU with and without historical occupancy reconstruction on the Occ3D-nuScenes.}
    \label{fig:valid_curve_histloss}
\end{minipage}
\end{table}

\noindent\textbf{Effect of historical occupancy reconstruction.}
We study the effect of historical occupancy reconstruction on the Occ3D-nuScenes validation set. As illustrated in Figure~\ref{fig:valid_curve_histloss}, incorporating the reconstruction loss consistently improves mIoU throughout training, indicating effective learning of spatio-temporal dynamics. The results demonstrate that jointly optimizing historical reconstruction and future prediction yields more coherent spatio-temporal representations, enhancing geometric fidelity and temporal consistency in long-horizon forecasting.
\section{Conclusion}
\label{sec:conclusion}
In this work, we present \shortname, a geometry-aware spatio-temporal context modeling method for 4D occupancy forecasting built on progressive explicit-implicit generation and dual-path spatio-temporal modeling. Different from the existing two-stage tokenize-then-generate paradigm, our method unifies historical scene reconstruction and future occupancy forecasting in a continuous BEV space. Specifically, the generation module ensures per-frame geometric consistency and semantic plausibility through pose-driven warping, motion-aware feature modulation, and attention-based feature refinement. The spatio-temporal module further enforces cross-frame coherence via global spatial context aggregation and temporal dynamics extraction, capturing long-range spatial dependencies and smooth scene evolution. The experimental results on the Occ3D-nuScenes demonstrate the superior effectiveness and efficiency of the proposed method. 

\section*{Acknowledgements}
This work was partially supported by the Joint Funds of the National Natural Science Foundation of China (Grant No.U24A20327), Guangdong S\&T Program under Grant 2026B0101110001.

\clearpage

%
%
\bibliographystyle{splncs04}
\bibliography{main}
\end{document}